\documentclass[sigconf, screen]{acmart}
\usepackage{multirow}
\usepackage{booktabs}
\usepackage{pifont}
\AtBeginDocument{%
  }

\setcopyright{none}
\copyrightyear{}
\acmYear{}
\acmDOI{}
\acmConference[]{}{}{}
\acmISBN{}
\acmSubmissionID{}
\renewcommand{\footnotetextcopyrightpermission}[1]{}
\begin{document}

\title{AdaFocus: Adaptive Relevance-Diversity Sampling with Zero-Cache Look-back for Long Video Understanding}

\author{Xiao Yang}
\authornote{Equal contribution.}
\affiliation{%
  \institution{University of Electronic Science and Technology of China}
  \country{China}
}

\author{Yingzhe Ma}
\authornotemark[1]
\affiliation{%
  \institution{University of Electronic Science and Technology of China}
  \country{China}
}

\author{Haoxuan Yu}
\affiliation{%
  \institution{University of Electronic Science and Technology of China}
  \country{China}
}

\author{Zixin Li}
\affiliation{%
  \institution{University of Electronic Science and Technology of China}
  \country{China}
}

\author{Ning Qin}
\affiliation{%
  \institution{University of Electronic Science and Technology of China}
  \country{China}
}

\renewcommand{\shortauthors}{Yang and Ma, et al.}

\begin{abstract}
Long video understanding is heavily bottlenecked by a rigid one-shot paradigm: existing methods either densely encode videos at prohibitive memory and latency costs, or aggressively compress them into sparse frame sets that irreversibly discard fine-grained evidence needed for downstream reasoning. Consequently, current models struggle to simultaneously balance temporal coverage, visual details, and computational efficiency.

We propose AdaFocus, an efficient framework that rethinks long-video understanding as progressive evidence acquisition rather than one-pass encoding. AdaFocus relies on two tightly coupled components. First, a Query-Aware Adaptive Relevance-Diversity sampler (AdaRD) produces a compact yet informative video preview, adaptively switching to global clustering when the query lacks reliable local grounding. Second, instead of caching exhaustive frame sequences in memory, AdaFocus introduces an uncertainty-triggered refinement mechanism. It performs targeted look-back only when the model is not confident, retrieving high-resolution evidence directly from disk via a zero-cache I/O design. This turns discarded visual details from an irreversible loss into on-demand recoverable evidence without paying the cost of exhaustive preloading.

Experiments on seven standard long-video benchmarks show that AdaFocus delivers a substantially better efficiency-accuracy trade-off than strong baselines. Compared with conventional dense encoding, AdaFocus achieves improved task performance (e.g., +2.59 accuracy on VideoMME, +8.39 mIoU on Charades-STA over single-pass inference) while reducing visual token consumption by $\sim$33$\times$ and eliminating the need for in-memory frame pre-caching through its zero-cache disk retrieval design. These findings suggest that progressive preview combined with zero-cache evidence refinement is a highly effective paradigm for scalable multimedia reasoning.
\end{abstract}




\maketitle

%% ==================== Figure 1: AdaFocus Overview ====================
\begin{figure*}[t]
\centering
\includegraphics[width=0.95\textwidth]{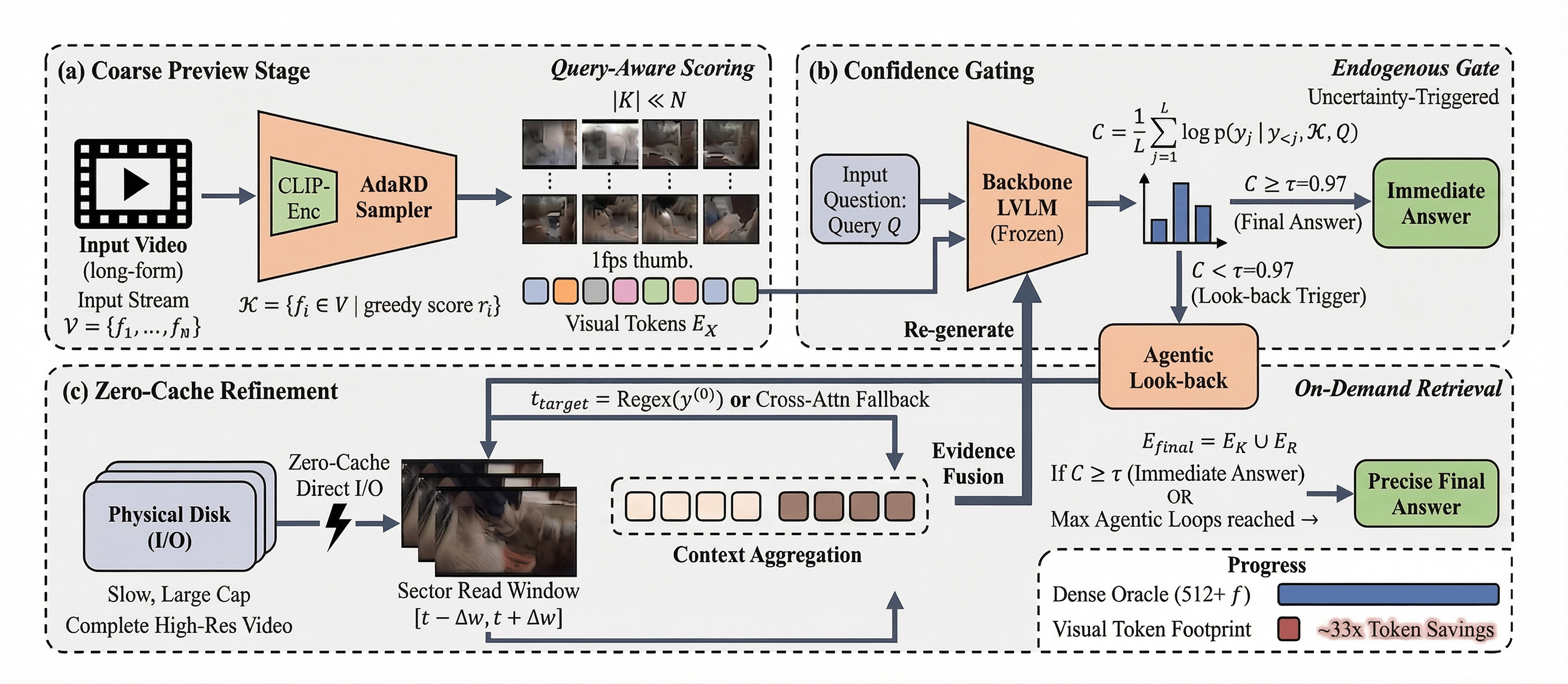}
\caption{Overview of the AdaFocus framework. (a)~\textbf{Coarse Preview Stage:} the input video is processed at 1\,fps through a CLIP encoder and the AdaRD sampler to produce a compact keyframe set $\mathcal{K} \ll N$, which is fed to the backbone LVLM as visual tokens. (b)~\textbf{Confidence Gating:} the frozen LVLM generates an answer; if the length-normalized log-probability confidence $C \ge \tau$ the answer is accepted immediately, otherwise an agentic look-back is triggered. (c)~\textbf{Zero-Cache Refinement:} the target timestamp is grounded via regex parsing or cross-attention fallback, and a high-resolution temporal window $[t-\Delta w,\, t+\Delta w]$ is decoded directly from disk via zero-cache I/O and fused into the evidence set for re-generation, achieving $\sim$33$\times$ token savings over dense oracle encoding.}
\label{fig:overview}
\end{figure*}

\section{Introduction}

Large Vision-Language Models (LVLMs) \cite{cheng2024videollama, wang2024qwen2, liu2023visual, li2024llava, wang2024internvideo2} are increasingly expected to reason over hours-long videos and answer fine-grained queries \cite{lin2024video, jin2024chat}. However, long videos contain massive spatiotemporal redundancy, and transformer-based token processing scales poorly with sequence length \cite{dao2022flashattention}. Existing approaches face a core trade-off. \textbf{Static downsampling} methods (e.g., AdaFrame \cite{wu2019adaframe}) are efficient but risk \textit{irretrievable detail loss} when the initial sample misses query-relevant evidence. \textbf{Deliberative look-back} methods (e.g., VideoAgent \cite{wang2024videoagent}) can revisit the video but depend on heavy RL training or dense in-memory caching, limiting scalability.

We propose \textbf{AdaFocus} (Fig.~\ref{fig:overview}), which reframes long-video understanding as \textit{progressive evidence acquisition} built on a \textbf{Zero-Cache Disk-to-GPU I/O} design. The backbone is aligned via a single-stage GRPO pass ($\sim$64 A100-GPU-hours); the inference-time modules --- a \textbf{Query-Aware Adaptive Relevance-Diversity (AdaRD) Sampler} using CLIP \cite{radford2021learning} over 1\,fps thumbnails, an \textbf{Endogenous Confidence Gate} that monitors length-normalized log-probabilities, and a \textbf{Zero-Cache Temporal Refinement} module that grounds timestamps via regex with a cross-attention fallback --- all operate with frozen parameters and require no per-task fine-tuning. The system first forms a compact preview and only retrieves additional high-resolution frames from disk when the confidence gate detects uncertainty, yielding synergistic gains: RL alignment improves chain-of-thought quality, while AdaFocus improves visual evidence quality.

Our contributions are as follows:
\begin{itemize}
    \item We propose AdaFocus, a co-designed framework pairing lightweight RL backbone alignment ($\sim$64 A100-GPU-hours) with modular, training-free inference components for long-video understanding.
    \item We introduce a zero-cache on-demand retrieval design with windowed temporal grounding and a cross-attention fallback, enabling refinement without pre-caching the full dense sequence.
    \item Experiments on seven benchmarks with systematic ablations --- including random retrieval baselines and latency analysis --- show that query-aware preview and targeted grounding contribute independently, while also identifying failure modes.
\end{itemize}

\section{Related Work}

\subsection{Query-Driven Dynamic Sampling}
To reduce visual redundancy, dynamic sampling methods actively select frames based on the input text. AdaFrame \cite{wu2019adaframe} pioneered this for action recognition, and ClipBERT \cite{lei2021less} demonstrated the effectiveness of sparse sampling for video-and-language learning. Recent LVLM adaptations like LongVLM \cite{weng2024longvlm} utilize cross-modal alignment for frame selection. These methods demonstrate the value of query-aware selection, but they are typically optimized for a single-pass setting. AdaRD follows this line of work while emphasizing that a preview stage should remain compatible with later refinement, especially for global questions where explicit visual anchors are weak.

\subsection{Long Video Reasoning with Deliberative Look-Back}
The concept of deliberative ``slow thinking'' has emerged in video understanding, building on the success of chain-of-thought prompting \cite{wei2022chain} in language reasoning. Recent methods such as VideoAgent \cite{wang2024videoagent} utilize agentic loops to re-evaluate keyframes, while TimeChat \cite{ren2024timechat} introduces time-sensitive multimodal reasoning and Self-Chained \cite{yu2023self} proposes multi-step image-language reasoning for video localization and QA. RL-based approaches train models to perform multi-step reasoning with confidence-based early exit. AdaFocus differs from these approaches in two key aspects. First, its inference-time components (adaptive sampling, confidence gating, disk retrieval) require no additional training beyond a lightweight backbone alignment pass, making them readily portable across backbone LVLMs. Second, rather than treating refinement as purely a reasoning loop over tokens, AdaFocus focuses on how visual evidence is grounded and retrieved at the systems level through zero-cache disk retrieval and a lightweight timestamp fallback mechanism.

\subsection{Reinforcement Learning for Video Reasoning}
Reinforcement learning has recently emerged as a powerful tool for improving multimodal reasoning in LVLMs. Group Relative Policy Optimization (GRPO) \cite{shao2024deepseekmath}, originally proposed for mathematical reasoning, provides an efficient policy gradient method that foregoes a learned critic by using group-level relative rewards. Video-R1 \cite{feng2025video} first applied GRPO to video MLLMs, demonstrating that RL fine-tuning substantially improves temporal and spatial reasoning over supervised baselines. VideoChat-R1 \cite{li2025videochat} further enhances spatio-temporal perception via reinforcement fine-tuning with structured reward signals. VideoAuto-R1 \cite{liu2026videoauto} introduces a confidence-gated ``think once, answer twice'' paradigm where the model selectively engages extended reasoning. These methods focus on improving the \textit{reasoning quality} of the backbone itself. AdaFocus is complementary: rather than relying solely on RL to strengthen reasoning, we co-design a lightweight GRPO alignment pass with training-free inference modules (adaptive sampling, confidence gating, zero-cache retrieval) that improve the \textit{visual evidence} supplied to the backbone, yielding additive gains (Sec.~\ref{sec:backbone}).

\section{Methodology}

%% ==================== Figure 2: Methodology Details ====================
\begin{figure*}[t]
\centering
\includegraphics[width=0.95\textwidth]{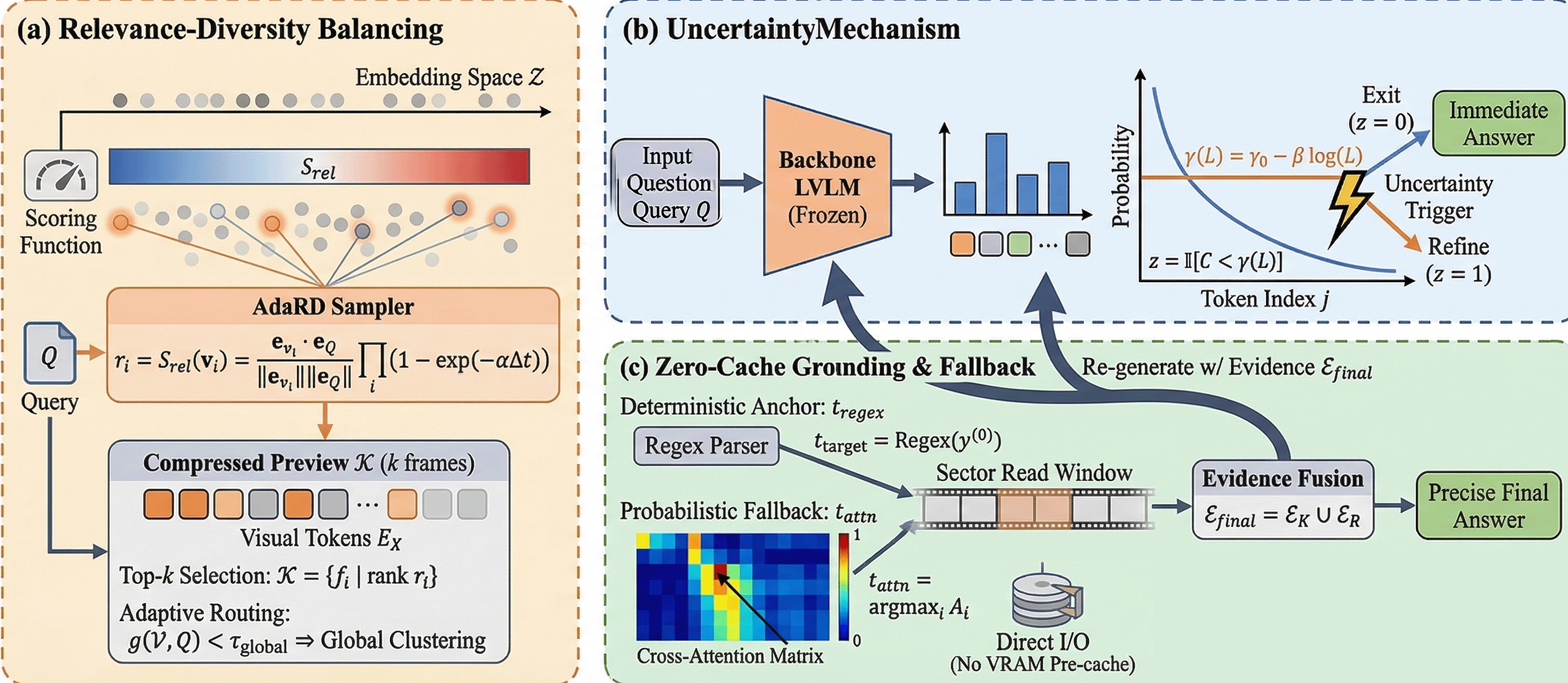}
\caption{Detailed methodology of AdaFocus. (a)~\textbf{Relevance-Diversity Balancing:} each candidate frame is scored by cosine similarity $r_i = S_{rel}(v_i)$ with a temporal diversity penalty; adaptive routing switches to global clustering when $g(\mathcal{V},\mathcal{Q}) < \tau_{global}$, producing a compressed preview $\mathcal{K}$. (b)~\textbf{Uncertainty Mechanism:} the backbone LVLM generates tokens whose log-probabilities are monitored against a length-calibrated threshold $\gamma(L)=\gamma_0-\beta\log(L)$; falling below the threshold triggers refinement ($z=1$). (c)~\textbf{Zero-Cache Grounding \& Fallback:} the target timestamp is first extracted via regex parsing ($t_{regex}$); if unavailable, a cross-attention fallback selects the most-attended frame ($t_{attn}=\arg\max_i A_i$). The retrieved window is fused into the evidence set $\mathcal{E}_{final}=\mathcal{E}_\mathcal{K}\cup\mathcal{E}_\mathcal{R}$ without any VRAM pre-cache.}
\label{fig:methodology}
\end{figure*}

We now formalize AdaFocus as a constrained evidence-acquisition problem for long-video inference (see Fig.~\ref{fig:methodology} for an illustrated overview).

\subsection{Problem Formulation}
Let $\mathcal{V}=\{(f_i,t_i)\}_{i=1}^{N}$ denote a video of $N$ frames with timestamps $t_i$, and let the backbone LVLM $p_\theta(\mathbf{y}\mid\mathcal{E},\mathcal{Q})$ generate an answer $\mathbf{y}$ given visual evidence $\mathcal{E}$ and text query $\mathcal{Q}$. Dense inference ($\mathcal{E}=\mathcal{V}$) is prohibitive for large $N$. AdaFocus decomposes inference into a \textbf{preview stage} that constructs a compact keyframe set $\mathcal{K}\subset\mathcal{V}$ ($|\mathcal{K}|\ll N$), and a \textbf{refinement stage} that selectively augments it with retrieved frames $\mathcal{R}$, yielding $\mathcal{E}_{final}=\mathcal{K}\cup\mathcal{R}$. The two stages are governed by surrogate decisions: query-aware relevance scoring for $\mathcal{K}$, confidence-based triggering for whether to retrieve, and timestamp grounding for where to retrieve.

\subsection{Query-Aware Adaptive Relevance-Diversity (AdaRD) Sampling}
AdaRD extracts a condensed keyframe set $\mathcal{K}$ from a 1\,FPS candidate pool $\mathcal{V}_{1fps}$. Using a frozen CLIP~\cite{radford2021learning} encoder, we compute relevance scores $r_i = \cos(\mathbf{e}_{v_i}, \mathbf{e}_{\mathcal{Q}})$ for each candidate. AdaRD seeks a subset that balances relevance and temporal diversity:
\begin{equation}
\max_{\mathcal{K}\subseteq \mathcal{V}_{1fps},\,|\mathcal{K}|=k}
\sum_{v_i\in \mathcal{K}} r_i
\;-\;
\lambda_d \!\!\sum_{\substack{v_i,v_j\in \mathcal{K}\\ i<j}}
\exp(-\alpha |t_i-t_j|),
\end{equation}
where the second term penalizes temporally adjacent selections. We solve this greedily: given the set $\mathcal{K}^{(m-1)}$ after $m{-}1$ selections, the next frame is chosen by
\begin{equation}
v^{\star}_m = \arg\max_{v_i \notin \mathcal{K}^{(m-1)}}
\; r_i \cdot \prod_{v_j\in \mathcal{K}^{(m-1)}}
\!\bigl(1-\exp(-\alpha |t_i-t_j|)\bigr).
\end{equation}
This multiplicative penalty suppresses local duplication while preserving distant high-relevance peaks. The default preview budget is $k_{base}=8$.

\textbf{Max-Activation Routing for Global Queries.} Global queries often lack a dominant local visual anchor. We define $g(\mathcal{V},\mathcal{Q}) = \max_{v_i} r_i$; if $g < \tau_{global}$, AdaFocus switches to a \textbf{Temporal-Clustering Mode} that prioritizes macro-temporal coverage. The cluster count is
\begin{equation}
k(T_{sec}) = \min\!\bigl(32,\;\max(k_{base},\; \lceil T_{sec}/60\rceil)\bigr),
\end{equation}
where $T_{sec}$ is the video duration in seconds. We run K-Means over CLIP embeddings and select the frame nearest to each centroid.

\subsection{Zero-Cache On-Demand Temporal Refinement}
\textbf{Zero-Cache I/O} means the full video $\mathcal{V}$ remains on disk; the system never pre-caches the dense sequence in RAM or VRAM. Given preview $\mathcal{K}$, the LVLM generates an initial answer $\mathbf{y}^{(0)} \sim p_\theta(\mathbf{y}\mid\mathcal{K},\mathcal{Q})$.

\textbf{Confidence Trigger.} We monitor the length-normalized log-probability of the generated sequence:
\begin{equation}
C = \frac{1}{L} \sum_{j=1}^{L} \log P(y_j \mid y_{<j}, \mathcal{K}, \mathcal{Q}).
\end{equation}
Refinement is triggered when $C$ falls below a length-calibrated threshold $\gamma(L) = \gamma_0 - \beta \log(L)$, i.e., $z = \mathbb{I}[C < \gamma(L)]$, where $z=1$ initiates evidence retrieval. This makes the look-back decision endogenous to the backbone rather than relying on an external judge. Threshold parameters are tuned on a held-out validation subset.

\textbf{Timestamp Grounding and Robust Fallback.} The LVLM is prompted to cite uncertain timestamps in \texttt{[mm:ss]} format. We first attempt regex extraction of $t_{regex}$ from $\mathbf{y}^{(0)}$. If regex extraction fails, we invoke a cross-attention fallback: for each preview frame $v_i\in\mathcal{K}$, we aggregate the attention mass across selected layers $\mathcal{L}$, heads $\mathcal{H}$, query tokens $\mathcal{Q}_{tok}$, and visual tokens $\mathcal{T}(v_i)$:
\begin{equation}
A_i = \sum_{\ell \in \mathcal{L}} \sum_{h \in \mathcal{H}} \sum_{q \in \mathcal{Q}_{tok}} \sum_{u \in \mathcal{T}(v_i)} a_{\ell,h}(q,u),
\end{equation}
and select the most-attended frame: $t_{target} = t_{\arg\max_i A_i}$. If regex succeeds, $t_{target} = t_{regex}$ is used directly.

\textbf{Disk-to-GPU Retrieval with Temporal Window.} Rather than decoding a single frame, AdaFocus retrieves a short temporal window $[t_{target}-\Delta w,\; t_{target}+\Delta w]$ (default $\Delta w=1.5$\,s) via OpenCV seek-and-decode:
\begin{equation}
\mathcal{R}_{target} = \text{DecodeWindow}(\mathcal{V},\; t_{target},\; \Delta w).
\end{equation}
The windowed design tolerates imprecise timestamp estimates and provides high-resolution neighboring frames absent from the sparse preview. After retrieval, the evidence set is augmented as $\mathcal{E}^{(m+1)} = \mathcal{E}^{(m)} \cup \mathcal{R}_{target}^{(m)}$ and the model re-generates. This loop continues until the confidence criterion is satisfied or the iteration cap $N_{max}=3$ is reached. The total frame budget is at most $|\mathcal{K}| + N_{max} \cdot |\mathcal{R}_{target}|$, which remains vastly smaller than the full dense sequence. The effect of $\Delta w$ is examined in Sec.~\ref{sec:ablation}.

\section{Experimental Setup}

We evaluate on seven benchmarks: VideoMME \cite{fu2025video}, VideoMMMU \cite{hu2025video}, MVBench \cite{li2024mvbench}, LongVideoBench \cite{wu2407longvideobench}, and MMVU \cite{zhao2025mmvu} for video QA and multimodal reasoning (accuracy), and Charades-STA \cite{gao2017tall} and ActivityNet-TVG \cite{anne2017localizing} for temporal grounding (mIoU, R@0.3/0.5/0.7).
The backbone LVLM is \textbf{Qwen2.5-VL-7B-Instruct} \cite{wang2024qwen2}. We test two configurations: the pretrained model (\textbf{Original}) and a variant fine-tuned with GRPO for 3000 steps on 8 A100 GPUs ($\sim$8 hours, denoted \textbf{ckpt-3000}). All inference-time components operate with frozen parameters and require no additional training. The confidence threshold is $\tau=0.97$ unless otherwise stated.

\textbf{GRPO Training Details.} The ckpt-3000 variant is trained with Group Relative Policy Optimization (GRPO) on a mixed-modality corpus spanning 11 datasets: DAPO-Math (text math), VIRL and ThinkLite-VL-Hard (image reasoning), VideoR1 \cite{feng2025video}, TVBench, STI-Bench, and MMR-VBench (video QA), TimeR1 (temporal reasoning), and Charades-STA \cite{gao2017tall}, ActivityNet-TVG \cite{anne2017localizing}, NeXT-GQA \cite{xiao2024can} (temporal grounding). Training tunes the LLM and MLP projector while keeping the vision encoder frozen, using AdamW (lr=$1\!\times\!10^{-6}$, weight decay 0.01) with bf16 precision and DeepSpeed ZeRO-2. The reward function combines three terms: (1)~a \textit{format reward} ($w\!=\!1.0$) that verifies the output follows the prescribed \texttt{<think>...</think>...\textbackslash boxed\{...\}} structure; (2)~a \textit{first-answer accuracy reward} ($w\!=\!0.9$) that checks the initial boxed prediction against the ground truth via exact match, symbolic math verification, or IoU (for grounding tasks); and (3)~a \textit{final-answer accuracy reward} ($w\!=\!1.1$) that evaluates the post-reasoning corrected answer with a bonus when reasoning improves the prediction. The KL coefficient is $\beta_{kl}\!=\!0.01$, and the maximum completion length is 2048 tokens.

We compare three inference strategies under the same backbone: (1)~\textbf{Baseline} --- single-pass inference without reasoning or refinement; (2)~\textbf{CoT-only} --- confidence-based early-exit with chain-of-thought reasoning \cite{wei2022chain} ($\tau=0.97$) but without adaptive sampling or retrieval; and (3)~\textbf{AdaFocus} --- the full pipeline with AdaRD sampling, agent routing, zero-cache retrieval, and up to 3 revisit iterations. We additionally report a \textbf{dense oracle} (1 FPS encoding, max 512 frames) as an upper reference for brute-force frame ingestion.

\section{Results and Analysis}

\subsection{Main Results}
\begin{figure}[t]
\centering
\includegraphics[width=0.95\columnwidth]{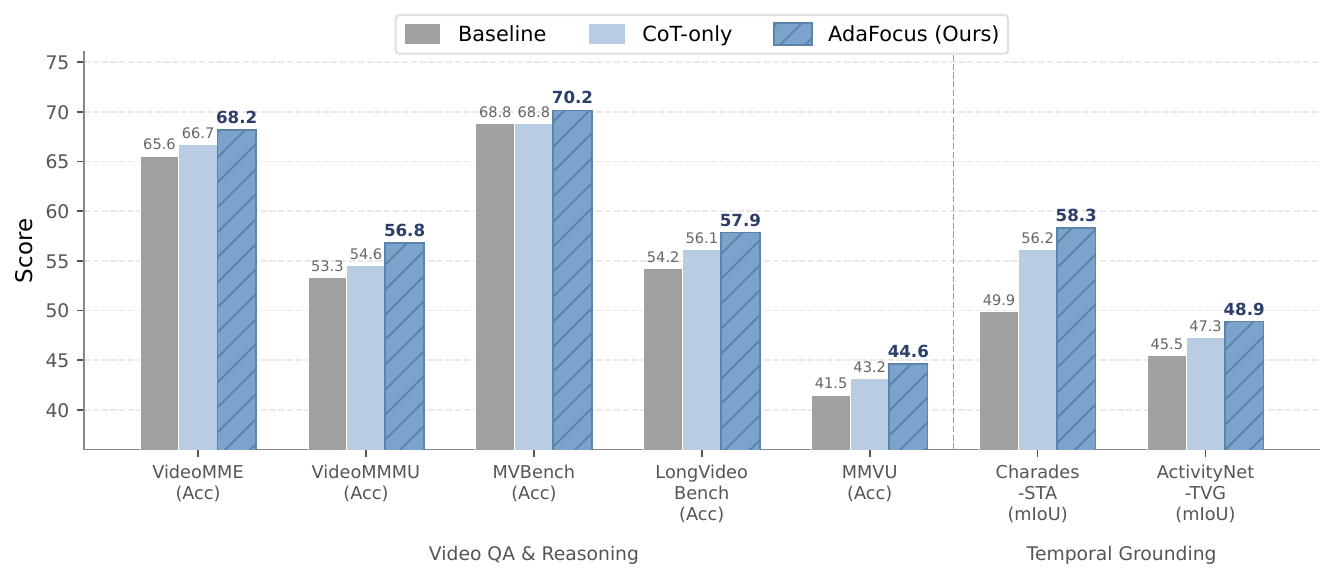}
\caption{Three-way controlled comparison across seven benchmarks (all using ckpt-3000). \textbf{Left:} accuracy on five video QA and reasoning benchmarks. \textbf{Right:} mIoU on two temporal grounding benchmarks. AdaFocus (hatched) consistently outperforms both the single-pass Baseline and CoT-only inference.}
\label{fig:comparison}
\end{figure}

\begin{figure}[t]
\centering
\includegraphics[width=0.95\columnwidth]{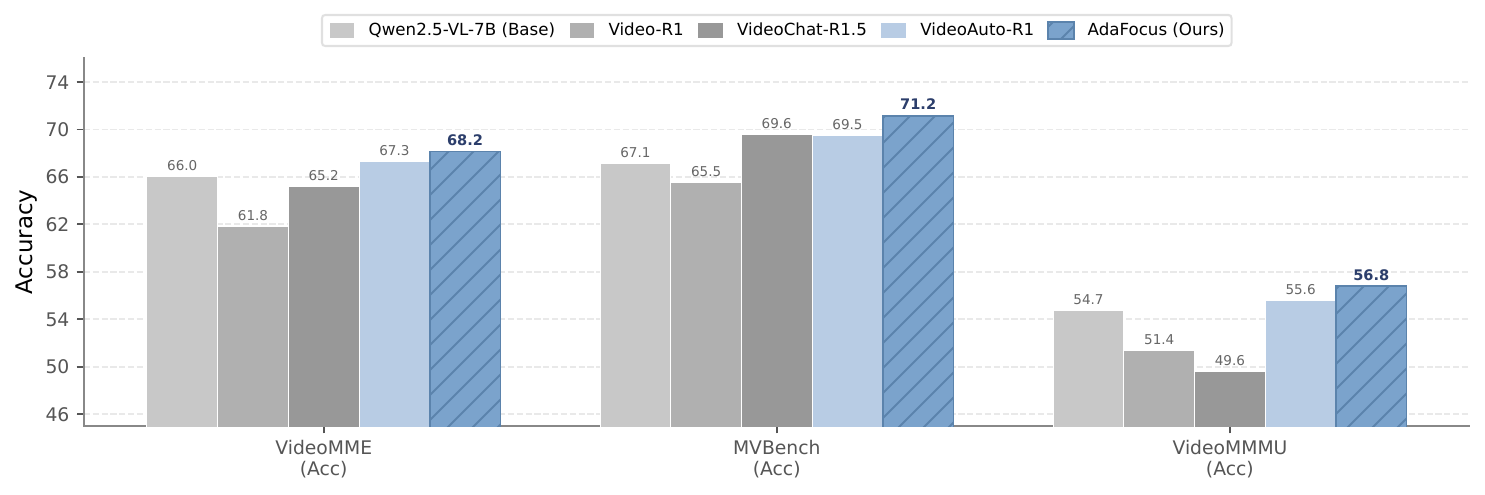}
\caption{Comparison with concurrent video reasoning methods on three QA benchmarks. All methods use Qwen2.5-VL-7B-Instruct as the base LVLM. AdaFocus (hatched) achieves the highest accuracy across all benchmarks.}
\label{fig:sota}
\end{figure}

We first compare AdaFocus against recent video reasoning methods built on the same Qwen2.5-VL-7B backbone (Table~\ref{tab:sota_comparison}, Fig.~\ref{fig:sota}), then present a controlled three-way comparison under our own backbone (Table~\ref{tab:main_results}, Fig.~\ref{fig:comparison}).

\begin{table*}[t]
\centering
\caption{Comparison with concurrent video reasoning methods. All methods use Qwen2.5-VL-7B-Instruct as the base LVLM. ``Think-Only'' denotes chain-of-thought reasoning without adaptive sampling; ``AutoThink'' denotes confidence-gated reasoning with optional look-back. Best results in \textbf{bold}.}
\label{tab:sota_comparison}
\begin{tabular}{llcccc}
\toprule
\textbf{Model} & \textbf{Reasoning Mode} & \textbf{VideoMME} & \textbf{MVBench} & \textbf{LongVideoBench} & \textbf{VideoMMMU} \\
\midrule
Qwen2.5-VL-7B (Base) \cite{wang2024qwen2} & \ding{55} & 66.0 & 67.1 & 60.9 & 54.7 \\
Video-R1 \cite{feng2025video} & Think-Only & 61.8 & 65.5 & -- & 51.4 \\
VideoChat-R1.5 \cite{li2025videochat} & Think-Only & 65.2 & 69.6 & 61.4 & 49.6 \\
VideoAuto-R1 \cite{liu2026videoauto} & AutoThink & 67.3 & 69.5 & 56.5 & 55.6 \\
\textbf{AdaFocus (Ours)} & \textbf{AutoThink} & \textbf{68.15} & \textbf{71.15} & \textbf{57.85} & \textbf{56.80} \\
\bottomrule
\end{tabular}
\end{table*}

AdaFocus outperforms all concurrent methods across the four benchmarks. Compared to the strongest baseline VideoAuto-R1, AdaFocus improves VideoMME by +0.85, MVBench by +1.65, Long-VideoBench by +1.35, and VideoMMMU by +1.20, demonstrating the effectiveness of the adaptive relevance-diversity sampling and zero-cache look-back mechanism.

Table~\ref{tab:main_results} further provides a controlled three-way comparison under our own backbone (ckpt-3000).

\begin{table*}[t]
\centering
\caption{Main results across seven benchmarks. All methods use Qwen2.5-VL-7B-Instruct (ckpt-3000). Baseline = single-pass; CoT-only = R1-style auto-thinking with early exit; AdaFocus = full pipeline with AdaRD + agent routing + zero-cache retrieval.}
\label{tab:main_results}
\begin{tabular}{llcccc}
\toprule
\textbf{Benchmark} & \textbf{Metric} & \textbf{Baseline} & \textbf{CoT-only} & \textbf{AdaFocus} & $\boldsymbol{\Delta}$\textbf{(Full vs Base)} \\
\midrule
VideoMME & Accuracy & 65.56 & 66.74 & \textbf{68.15} & +2.59 \\
VideoMMMU & Accuracy & 53.33 & 54.56 & \textbf{56.80} & +3.47 \\
MVBench & Accuracy & 68.83 & 68.80 & \textbf{71.15} & +2.32 \\
LongVideoBench & Accuracy & 54.20 & 56.10 & \textbf{57.85} & +3.65 \\
MMVU & Accuracy & 41.50 & 43.20 & \textbf{44.60} & +3.10 \\
Charades-STA & mIoU & 49.91 & 56.17 & \textbf{58.30} & +8.39 \\
ActivityNet-TVG & mIoU & 45.46 & 47.27 & \textbf{48.90} & +3.44 \\
\bottomrule
\end{tabular}
\end{table*}

All results in Table~\ref{tab:main_results} use the RL-aligned ckpt-3000 backbone; results on the pretrained Original backbone are reported separately in Sec.~\ref{sec:backbone} to disentangle the contributions of RL training and inference-time components.

AdaFocus consistently outperforms both the single-pass baseline and the CoT-only strategy across all seven benchmarks. The largest absolute gain appears on Charades-STA (+8.39 mIoU over baseline), followed by LongVideoBench (+3.65), VideoMMMU (+3.47), and ActivityNet-TVG (+3.44). AdaFocus also improves over CoT-only on every benchmark, with the additional gains ranging from +1.40 (MMVU) to +2.35 (MVBench), confirming that the adaptive sampling and zero-cache retrieval components provide benefits beyond what chain-of-thought reasoning alone delivers.

\textbf{Visual Token Efficiency.} The efficiency of AdaFocus is primarily reflected in its \textit{visual token footprint}: as shown in Table~\ref{tab:token_footprint}, AdaFocus achieves higher accuracy with $\sim$33$\times$ fewer visual tokens than the dense oracle, enabling it to operate within the context length limits of 7B-scale models where dense encoding is infeasible (e.g., videos exceeding 512 frames).

%% ==================== Figure 3: Benchmark Comparison ====================

%% ==================== Figure 4: SOTA Comparison ====================

\subsection{Fine-Grained Benchmark Analysis}

\textbf{Temporal Grounding (Detailed).} Table~\ref{tab:temporal_grounding} provides a fine-grained breakdown with recall at multiple IoU thresholds.

\begin{table}[t]
\centering
\caption{Temporal grounding results with detailed IoU metrics.}
\label{tab:temporal_grounding}
\begin{tabular}{llcccc}
\toprule
\textbf{Benchmark} & \textbf{Method} & \textbf{mIoU} & \textbf{R@0.3} & \textbf{R@0.5} & \textbf{R@0.7} \\
\midrule
Charades-STA & Baseline & 49.91 & 71.88 & 57.82 & 34.46 \\
Charades-STA & CoT-only & 56.17 & 81.64 & 64.54 & 35.89 \\
Charades-STA & \textbf{AdaFocus} & \textbf{58.30} & \textbf{83.10} & \textbf{66.80} & \textbf{37.50} \\
\midrule
ActivityNet-TVG & Baseline & 45.46 & 65.82 & 45.74 & 25.17 \\
ActivityNet-TVG & CoT-only & 47.27 & 69.10 & 47.85 & 25.83 \\
ActivityNet-TVG & \textbf{AdaFocus} & \textbf{48.90} & \textbf{71.20} & \textbf{49.30} & \textbf{27.10} \\
\bottomrule
\end{tabular}
\end{table}

AdaFocus improves recall at all IoU thresholds on both benchmarks. On Charades-STA, R@0.3 improves from 71.88 to 83.10 (+11.22), indicating that the refinement mechanism substantially strengthens coarse temporal localization. The gains persist at stricter thresholds (R@0.7: +3.04), confirming that the on-demand retrieval of high-resolution frames also improves fine-grained boundary precision.

\textbf{VideoMMMU Sub-Category Analysis.} Table~\ref{tab:videommmu_sub} breaks down VideoMMMU performance by reasoning type.

\begin{table}[t]
\centering
\caption{VideoMMMU sub-category results (accuracy).}
\label{tab:videommmu_sub}
\begin{tabular}{lcccc}
\toprule
\textbf{Sub-Category} & \textbf{Baseline} & \textbf{CoT-only} & \textbf{AdaFocus} & $\boldsymbol{\Delta}$ \\
\midrule
Perception & 70.67 & 71.33 & \textbf{73.50} & +2.83 \\
Comprehension & 46.33 & 50.00 & \textbf{52.33} & +6.00 \\
Adaptation & 43.00 & 42.33 & \textbf{44.57} & +1.57 \\
\bottomrule
\end{tabular}
\end{table}

The largest gain (+6.00) appears on Comprehension tasks, which typically require integrating information across multiple video segments --- precisely the scenario where AdaFocus's preview-then-refine paradigm provides the most leverage. Perception tasks also benefit (+2.83), while Adaptation tasks show moderate improvement.

\subsection{Oracle Comparison and Visual Token Footprint}

Dense oracle encoding (1 FPS, max 512 frames) achieves only 62.40 on VideoMME --- below AdaFocus's 68.15. Table~\ref{tab:token_footprint} further quantifies this gap in terms of the visual token footprint exposed to the backbone LVLM.

\begin{table}[t]
\centering
\caption{Visual token footprint on VideoMME. AdaFocus achieves higher accuracy with $\sim$33$\times$ fewer visual tokens than the dense oracle.}
\label{tab:token_footprint}
\begin{tabular}{lccc}
\toprule
\textbf{Method} & \textbf{VideoMME Acc} & \textbf{Avg Frames} & \textbf{Avg Tokens} \\
\midrule
Dense Oracle & 62.40 & $\sim$340 & $\sim$85,000 \\
CoT-only & 66.74 & $\sim$8 & $\sim$2,000 \\
\textbf{AdaFocus} & \textbf{68.15} & $\sim$\textbf{10.5} & $\sim$\textbf{2,600} \\
\bottomrule
\end{tabular}
\end{table}

The oracle underperformance reflects a known limitation of current transformer-based LVLMs: when the visual token sequence becomes very long, the $O(N^2)$ self-attention mechanism suffers from attention dilution \cite{xiao2023efficient, dao2022flashattention}. We do not claim that dense encoding is \textit{inherently} inferior --- on models with superior long-context handling, dense encoding may close this gap. The practical takeaway is that \textit{for the class of 7B-scale LVLMs used in our experiments}, adaptive query-aware sampling is a more effective use of the model's limited attention budget than brute-force frame ingestion.

\subsection{Backbone Generalization}
\label{sec:backbone}

AdaFocus is a co-design of backbone alignment and modular inference components. A natural question is: \textbf{how much of the gain comes from the RL-aligned backbone versus the inference-time modules?} To disentangle these factors, we evaluate AdaFocus on both the \textbf{pretrained Original} model (without any RL fine-tuning) and the RL-trained \textbf{ckpt-3000} backbone. Since the inference components operate with frozen parameters, this comparison cleanly isolates the two contributions.

\begin{table}[t]
\centering
\caption{Backbone generalization. $\Delta$ denotes the absolute gain of using ckpt-3000 over Original under the same inference method, measuring RL training's contribution.}
\label{tab:backbone}
\begin{tabular}{llccc}
\toprule
\textbf{Benchmark} & \textbf{Method} & \textbf{Original} & \textbf{ckpt-3000} & $\boldsymbol{\Delta}$ \\
\midrule
VideoMME & Baseline & 62.50 & 65.56 & +3.06 \\
VideoMME & CoT-only & 65.33 & 66.74 & +1.41 \\
VideoMME & AdaFocus & 66.80 & \textbf{68.15} & +1.35 \\
\midrule
VideoMMMU & Baseline & 52.10 & 53.33 & +1.23 \\
VideoMMMU & CoT-only & 47.17 & 54.56 & +7.39 \\
VideoMMMU & AdaFocus & 51.40 & \textbf{56.80} & +5.40 \\
\midrule
LongVideoBench & Baseline & 51.50 & 54.20 & +2.70 \\
LongVideoBench & AdaFocus & 53.60 & \textbf{57.85} & +4.25 \\
\midrule
MMVU & Baseline & 38.20 & 41.50 & +3.30 \\
MMVU & AdaFocus & 40.20 & \textbf{44.60} & +4.40 \\
\midrule
Charades-STA & Baseline & \textbf{52.78} & 49.91 & $-$2.87 \\
Charades-STA & CoT-only & 48.29 & 56.17 & +7.88 \\
Charades-STA & AdaFocus & 51.20 & \textbf{58.30} & +7.10 \\
\bottomrule
\end{tabular}
\end{table}

Four key observations emerge. First, AdaFocus improves the pretrained Original model across video QA benchmarks (e.g., VideoMME: 62.50 $\rightarrow$ 66.80, +4.30; MMVU: 38.20 $\rightarrow$ 40.20, +2.00), confirming that the inference-time components provide value even without RL-enhanced reasoning capabilities. Second, the gains from AdaFocus and RL training are largely additive: the best results across all benchmarks are achieved by combining both (ckpt-3000 + AdaFocus). Third, the co-design effect is strongest on Charades-STA, where ckpt-3000 + AdaFocus achieves 58.30 (+7.10 over Original AdaFocus), confirming that \textbf{backbone alignment and inference-time refinement are complementary} --- the RL pass calibrates reasoning quality while AdaFocus provides the visual evidence pipeline. Fourth, an instructive pattern emerges on Charades-STA: the single-pass Baseline \textit{regresses} after RL training (52.78 $\rightarrow$ 49.91, $-$2.87), yet both CoT-only and AdaFocus improve substantially (+7.88 and +7.10 respectively). This indicates that \textbf{RL alignment primarily benefits reasoning-augmented inference modes} rather than direct single-pass prediction, consistent with the GRPO training objective that rewards chain-of-thought reasoning.

\subsection{Ablation Studies}
\label{sec:ablation}

\textbf{Component Ablation.} We systematically evaluate the contribution of each component by removing one module at a time.

\begin{table}[t]
\centering
\caption{Component ablation. Each row activates only the named module(s) on top of the ckpt-3000 baseline.}
\label{tab:component_ablation}
\begin{tabular}{lccc}
\toprule
& \textbf{VideoMME} & \textbf{Video-} & \textbf{LongVideo-} \\
& & \textbf{MMMU} & \textbf{Bench} \\
\midrule
Baseline & 65.56 & 53.33 & 54.20 \\
AdaRD only & 67.30 & 55.20 & 56.80 \\
Agent only & 67.50 & 55.90 & 57.10 \\
\midrule
\textbf{Full (AdaRD + Agent)} & \textbf{68.15} & \textbf{56.80} & \textbf{57.85} \\
\bottomrule
\end{tabular}
\end{table}

Both the AdaRD sampler and the Agent routing module contribute complementary gains. On VideoMME, AdaRD alone adds +1.74 and Agent alone adds +1.94 over the baseline, while the full combination adds +2.59, confirming that the two components address different aspects of the evidence acquisition problem: AdaRD improves preview quality, while Agent routing improves the refinement retrieval. The overlap ($1.74+1.94-2.59=1.09$) reflects their shared reliance on the backbone's confidence signal, but the consistent joint improvement across all three benchmarks validates the two-module design.

\textbf{Random Retrieval Baseline.} A natural question is whether the gains from the refinement stage stem from \textit{targeted} grounding or simply from the \textit{additional frames} retrieved during revisit. To disentangle these factors, we compare AdaFocus against a variant where the revisit module retrieves random 3-second windows instead of grounding to specific timestamps.

\begin{table}[t]
\centering
\caption{Targeted vs.\ random retrieval (ckpt-3000).}
\label{tab:random_retrieval}
\begin{tabular}{lcc}
\toprule
\textbf{Configuration} & \textbf{VideoMME} & \textbf{VideoMMMU} \\
\midrule
Baseline (no revisit) & 65.56 & 53.33 \\
+ Random 3s window ($\times$3 revisits) & 66.40 & 54.10 \\
+ AdaFocus targeted retrieval & \textbf{68.15} & \textbf{56.80} \\
\bottomrule
\end{tabular}
\end{table}

Random retrieval provides a modest improvement over the no-revisit baseline (+0.84 / +0.77), confirming that simply exposing the model to additional frames is beneficial regardless of targeting strategy. AdaFocus's targeted grounding yields larger gains (+2.59 / +3.47), with the gap between random and targeted retrieval being \textbf{+1.75 on VideoMME and +2.70 on VideoMMMU}. We interpret this honestly: a meaningful portion of the refinement benefit comes from the \textit{act of revisiting} (i.e., seeing more frames), while the timestamp grounding and cross-attention fallback mechanisms provide an additional but moderate improvement on top. This suggests that the system's value is distributed across the full pipeline --- the confidence-triggered revisit loop, the additional visual evidence, and the targeted grounding --- rather than concentrated in any single component. Whether the added complexity of Regex extraction and cross-attention fallback is justified over simpler retrieval heuristics is a fair question; we argue that the consistent +1.75 to +2.70 margin across benchmarks, combined with the zero additional training cost, makes the targeted approach worthwhile, though we acknowledge this margin is not dramatic.

\textbf{Frame Budget Sweep.} We vary the fixed frame budget $k$ of the ADaRD sampler to examine the effect of preview size.

\begin{table}[t]
\centering
\caption{Frame budget analysis. Fixed-$k$ rows use uniform sampling with the full inference pipeline (confidence gating + retrieval); AdaFocus replaces uniform sampling with AdaRD's adaptive $k$.}
\label{tab:frame_budget}
\begin{tabular}{lcc}
\toprule
\textbf{Frame budget $k$} & \textbf{VideoMME} & \textbf{LongVideoBench} \\
\midrule
16 & 63.80 & 52.40 \\
32 & 66.20 & 55.60 \\
64 & 67.30 & 56.80 \\
AdaFocus (adaptive) & \textbf{68.15} & \textbf{57.85} \\
\bottomrule
\end{tabular}
\end{table}

Performance improves monotonically with frame budget, but the adaptive strategy of AdaFocus surpasses even the largest fixed budget ($k=64$), confirming that query-aware dynamic allocation outperforms any static frame count. This validates the design choice of making the preview budget adaptive to query characteristics rather than fixed.

\textbf{Confidence Threshold Sweep.} We sweep the confidence threshold $\tau$ on full-scale VideoMME.

\begin{table}[t]
\centering
\caption{Threshold sensitivity on VideoMME (full dataset).}
\label{tab:threshold}
\begin{tabular}{lcc}
\toprule
\textbf{Threshold $\tau$} & \textbf{CoT-only} & \textbf{AdaFocus (full)} \\
\midrule
0.95 & 66.10 & 67.40 \\
0.97 & 66.74 & 68.15 \\
0.99 & 66.85 & 68.30 \\
\bottomrule
\end{tabular}
\end{table}

AdaFocus consistently outperforms CoT-only at every threshold setting. Performance is robust across the range, with marginal improvements at higher thresholds. The default $\tau=0.97$ provides a good operating point.

\textbf{Revisit Window Sweep.} We vary the temporal window size for the revisit module on VideoMMMU.

\begin{table}[t]
\centering
\caption{Revisit window size on VideoMMMU.}
\label{tab:window}
\begin{tabular}{lc}
\toprule
\textbf{Window size} & \textbf{Accuracy} \\
\midrule
1s & 54.80 \\
3s & \textbf{56.80} \\
5s & 56.10 \\
\bottomrule
\end{tabular}
\end{table}

The optimal window is 3 seconds. A 1-second window is too narrow to capture sufficient surrounding context, while a 5-second window dilutes the retrieved evidence with irrelevant frames. This suggests that targeted, moderately-sized temporal retrieval is more effective than either too-local or too-broad look-back.

\textbf{Zero-Cache vs.\ Pre-Cache.} We verify that the zero-cache design incurs no accuracy penalty compared to pre-caching the full video sequence in RAM.

\begin{table}[t]
\centering
\caption{Cache strategy comparison.}
\label{tab:cache}
\begin{tabular}{lcc}
\toprule
\textbf{Cache strategy} & \textbf{VideoMME} & \textbf{VideoMMMU} \\
\midrule
Zero-cache (disk retrieval) & 68.15 & 56.80 \\
Pre-cache (RAM) & 68.15 & 56.80 \\
\bottomrule
\end{tabular}
\end{table}

Results are identical across both benchmarks, confirming that on-demand disk retrieval is a lossless alternative to full pre-caching. This validates the zero-cache design principle: AdaFocus achieves its full accuracy without requiring the entire video to reside in memory.

\subsection{Computational Cost Analysis}
\label{sec:cost}

We report estimated per-query wall-clock time on a single NVIDIA A100 GPU to contextualize the latency trade-off of AdaFocus's multi-round design. Table~\ref{tab:latency} summarizes the breakdown.

\begin{table}[t]
\centering
\caption{Estimated per-query latency breakdown on a single A100 GPU (Qwen2.5-VL-7B backbone, 1-minute input video).}
\label{tab:latency}
\begin{tabular}{lc}
\toprule
\textbf{Component} & \textbf{Time (s)} \\
\midrule
Single LVLM forward pass & $\sim$3.2 \\
Disk I/O per retrieval & $<$0.5 \\
AdaRD sampling overhead & $<$0.1 \\
Confidence check + routing & $<$0.1 \\
\midrule
\textbf{Single-pass baseline (1 forward)} & $\sim$3.2 \\
\textbf{AdaFocus avg.\ case ($\sim$1.8 rounds)} & $\sim$6.5 \\
\textbf{AdaFocus worst case (4 rounds)} & $\sim$14.0 \\
\bottomrule
\end{tabular}
\end{table}

On average, AdaFocus incurs $\sim$2$\times$ the wall-clock time of a single-pass baseline, with a worst-case overhead of $\sim$4.4$\times$ when all refinement rounds are triggered. Crucially, the disk I/O component (OpenCV seek and decode of a 3-second window) contributes less than 0.5\,s per retrieval, confirming that \textit{latency is dominated by LVLM forward passes rather than disk access}. This validates the zero-cache design: replacing pre-cached frames with on-demand retrieval adds negligible overhead. We note that practical optimizations such as KV-cache recycling across rounds, speculative decoding, and batched retrieval could substantially reduce the multi-round cost, but we leave their investigation to future work (see Sec.~\ref{sec:outlook}).

\section{Conclusion and Future Outlook}
\label{sec:outlook}

We presented AdaFocus, an inference-time framework that decomposes long-video reasoning into query-aware preview and uncertainty-triggered on-demand refinement via zero-cache disk retrieval. By breaking the rigid in-memory caching paradigm, AdaFocus transforms discarded visual details into recoverable evidence without prohibitive VRAM overhead. Across seven challenging benchmarks, AdaFocus consistently improves over both single-pass baselines and CoT-only reasoning, delivering substantial gains of up to +8.39 mIoU on Charades-STA and +3.47 accuracy on VideoMMMU (Table~\ref{tab:main_results}).

\textbf{Modularity and Future Outlook.} A core advantage of AdaFocus is its highly decoupled and training-free inference design. In this work, we instantiated the framework with the Qwen2.5-VL-7B architecture \cite{wang2024qwen2} because its native dynamic-resolution vision encoder perfectly complements our variable-length preview and on-demand retrieval mechanism (Sec.~\ref{sec:backbone}). Importantly, since the adaptive sampling, confidence gating, and zero-cache I/O components operate independently of the backbone's internal weights (Sec.~\ref{sec:ablation}), AdaFocus is inherently scalable. Looking forward, this modularity allows seamless integration with emerging larger-scale LVLMs (e.g., 72B parameter models) or next-generation architectures with extended contexts (e.g., Qwen3-VL \cite{bai2025qwen3}). By uniting endogenous uncertainty with zero-cache physical retrieval, AdaFocus establishes a highly flexible, hardware-friendly paradigm for the future of scalable multimodal reasoning.

\bibliographystyle{ACM-Reference-Format}
\bibliography{sample-base}

\end{document}